\pdfoutput=1
\documentclass[10pt,twocolumn,letterpaper]{article}

\usepackage{iccv}
\usepackage{times}
\usepackage{epsfig}
\usepackage{graphicx}
\usepackage{amsmath}
\usepackage{amssymb}

\usepackage{enumitem}
\usepackage{verbatim}

\usepackage{epsf,epstopdf}
\usepackage{subfigure}
\def\degree{${}^{\circ}$}

\usepackage[pagebackref=true,breaklinks=true,letterpaper=true,colorlinks,bookmarks=false]{hyperref}

 \iccvfinalcopy 

\def\iccvPaperID{874} 

\def\be{\begin{equation}}
\def\ee{\end{equation}}
\def\bea{\begin{eqnarray*}}
\def\eea{\end{eqnarray*}}
\ificcvfinal\fi

\begin{document}

\title{Flip-Rotate-Pooling Convolution and Split Dropout on Convolution Neural Networks for Image Classification}


\author{{\bf Fa Wu \quad Peijun Hu  \quad Dexing Kong}\\
\medskip
{School of Mathematical Sciences, Zhejiang University}\\
\medskip
{\small \{wufa85, \, peijunhu, \, dkong\}@zju.edu.cn}
}

\maketitle

\begin{abstract}
    This paper presents a new version of Dropout called Split Dropout (sDropout) and rotational convolution techniques to improve CNNs' performance on image classification. The widely used standard Dropout has advantage of preventing deep neural networks from overfitting by randomly dropping units during training. Our sDropout randomly splits the data into two subsets and keeps both rather than discards one subset. We also introduce two rotational convolution techniques, i.e. rotate-pooling convolution (RPC) and flip-rotate-pooling convolution (FRPC) to boost CNNs' performance on the robustness for rotation transformation. These two techniques encode rotation invariance into the network without adding extra parameters. Experimental evaluations on ImageNet2012 classification task demonstrate that sDropout not only enhances the performance but also converges faster. Additionally, RPC and FRPC make CNNs more robust for rotation transformations. Overall, FRPC together with sDropout bring $1.18\%$ (model of Zeiler and Fergus~\cite{zeiler2013visualizing}, 10-view, top-1) accuracy increase in ImageNet 2012 classification task compared to the original network.
\end{abstract}

\section{Introduction}

    Since proposed by LeCun \etal \cite{lecun1989backpropagation} in the early 1990’s, convolutional neural networks (CNNs) have demonstrated excellent performance in visual tasks, such as image classification and object recognition. Recently, deep convolutional neural networks have shown state-of-the-art performance in challenging large-scale image classification. Krizhevsky \etal~\cite{krizhevsky2012imagenet} show record beating performance on the ImageNet 2012 classification benchmark ~\cite{Russakovsky14}, with their convnet model achieving an error rate of $16.4\%$, compared to the 2nd place result of $26.1\%$. The recent work of He \etal~\cite{he2015delving} shows an classification accuracy exceeding what human can achieve for the first time.

    Several factors contribute to the success of deep CNNs in image classification: availability of large-scale labeled training data, powerful GPU support and effective regularization strategies. When large datasets are available, CNNs are capable to extract effective features from the images by designing more layers and adding units to the networks. However, overfitting is really a problem in such networks with a large number of parameters. Fortunately, a wide range of techniques for regularizing have been developed, such as adding an $l_2$ penalty on the network weights, Bayesian methods, weights elimination and early stop of training. Among the regularizing techniques, the Dropout proposed by Hinton \etal~\cite{hinton2012improving} is an effective way to not only reduce overfitting, but also give great improvements on many benchmark tasks.

    Additionally, the two key concepts of CNNs, i.e. local receptive fields and weight-tying, help the networks encode translational invariance into the features learned from images. CNNs take translated versions of the basis function (the convolution filters) and ”pool” over them. In this way, different image locations share the same basis function and thus the number of parameters to be learned is significantly reduced. By pooling over neighboring units CNNs hard-code translation invariance into the model. However, this strategy prevents the pooling units from capturing more complex invariance, such as rotation invariance. The experiments of Zeiler and Fergus~\cite{zeiler2013visualizing} validate that the outputs of CNNs are not invariant to rotation transformation.

    In this paper, we improve the CNNs in two aspects: a new version of Dropout and the improvement of rotation invariance of CNNs. The core of standard Dropout proposed by Hinton \etal~\cite{hinton2012improving} is to randomly drop half the units in each fully-connected layer in forward propagation and then take the other half of the units through back propagation. This prevents units from co-adapting too much, but only half of the parameters are trained in every training iteration. Here, we propose a new version, called Split Dropout (sDropout), which splits the units into two subsets rather than drops half of them. In this way, all the weights are trained while the co-adapting between units is broken. sDropout can be a substitute for Dropout in any circumstances. To add rotation invariance into the model, we introduce two rotational convolution techniques, i.e. rotated-pooling convolution (RPC) and flip-rotate-pooling convolution (FRPC), to the convolution layers of CNNs. By rotating or flipping the convolution filters, and then through convolution and max-pooling, the modified CNNs are more capable to extract rotational/flipping invariant features.

    We validate our methods on the large-scale 1000-class ImageNet 2012 classification datasets \cite{Russakovsky14}. Our model is based on the architecture of Zeiler and Fergus~\cite{zeiler2013visualizing} with Rectified Linear Unit (ReLU)~\cite{nair2010rectified} and Parametric Rectified Linear Unit (PReLU)~\cite{he2015delving} respectively. By replacing the Dropout with sDropout, we observe not only lower testing error but also faster convergence in training. Additionally, our RPC and FRPC do actually improve the classification accuracy and demonstrate rotation invariance on rotated images. We note that sDropout and rotational convolution techniques do not change the original architecture and add very little memory and computation cost.


    \section{\label{section:related work}Related Work}

    In the last few years, we have witnessed tremendous performance improvements made on the CNNs. These improvements are mainly two types: building more powerful architectures and designing effective regularizing techniques. To be more capable of fitting training datasets, especially for large-scale datasets, models are made deeper and larger, e.g., the work of Simonyan \etal~\cite{simonyan2014very} and Szegedy \etal~\cite{szegedy2014going}. Setting strides smaller to capture more image information also helps in model improvement , such as the work of Zeiler and Fergus ~\cite{zeiler2013visualizing}. Rectified Linear Unit (ReLU) ~\cite{nair2010rectified} and Parametric Rectified Linear Unit (PReLU)~\cite{he2015delving} contribute to the recent success of improvements on activation functions. On the other hand, regularization is an important aspect in boosting the testing performance of CNNs. Data augmentation ~\cite{krizhevsky2012imagenet, howard2013some} and the Dropout ~\cite{hinton2012improving} technique are recently the common way to regularize the model. One major work in this paper is to improve Dropout.

    In standard Dropout ~\cite{Hinton12}, each element of a layer's output is kept with probability p, otherwise set to 0 with probability (1 - p) (usually set p = 0.5). It can be seen as a stochastic regularization technique. Since the successful application of Dropout in feedforward neural networks for speech and object recognition ~\cite{Hinton12}, several works have been done on the improvement and analysis of this technique. A generalization of Dropout, 'standout' proposed by Ba \etal~\cite{ba2013adaptive}, uses a binary belief network to compute the probability for each hidden variable. They believe several hidden units are highly correlated in the pre-dropout activities. DropConnect, proposed by Wan \etal~\cite{wan2013regularization}, sets a randomly selected subset of weights, rather than activations, to zero. To speed up the training process in Dropout, Wang \etal~\cite{wang2013fast} propose a fast dropout. The model uses an objective function approximately equivalent to that of real standard dropout training, but does not actually sample the inputs. sDropout is an extension of Dropout and can be easily combined with methods mentioned above without conflict.

    In spite of recent progresses in feature extraction from images~\cite{bengio2013representation}, representing complex invariance of images in different learning systems is still a challenging task and attract many works. In unsupervised deep learning systems, Zou \etal ~\cite{zou2012deep} utilize slowness and non-degeneracy principle which can also be applied to still images. Le ~\cite{le2013building} proposes an autoencoder to recognize faces. The feature detector is reported to be robust to translation. Zeiler \etal~\cite{zeiler2011adaptive} present a hierarchical model using alternating layers of convolutional sparse coding and 3D max-pooling to learn image decompositions. In object recognition, a new image is decomposed into multiple layers of features by the learned model and then recognized by a classifier. Ngiam \etal ~\cite{ngiam2010tiled} propose a tiled convolutional neural network that does not require adjacent hidden units to share identical weights, but only needs hidden units $k$ steps away from each other to have tiled weights. With TICA pretraining strategy, the architecture is able to learn scale and rotation invariance. Wavelet scatting networks, introduced in \cite{mallat2010recursive, mallat2012group} build translation invariant image representations with average poolings of wavelet modulus coefficients. The follow-on work by Sifre \etal \cite{sifre2013rotation} propose a joint translation and rotation invariant representation of image patches, which is implemented with a deep convolution network. The filters of the deep convolution network are not learned but are scaled and rotated wavelets.

    A more direct way to add invariance is Data augmentation. Related work of Sermanet \etal \cite{sermanet2011traffic} build a jittered dataset by adding transformed versions of the original training set, including translation, scaling and rotation. There are many more works that transform input data to obtain invariance, e.g. Howard and Andrew G \cite{howard2013some}, Dosovitskiy \etal \cite{dosovitskiy2013unsupervised}. But augmenting the training set with rotated versions does not achieve the same effect as ours, as it can not be restricted to upper layers. The rotation invariance in our work is achieved by pooling over systematically transformed versions of filters. It is closely related to the recent work of Gens \etal \cite{gens2014deep}, which pools features over symmetry groups within a neural network.

\section{\label{section:approach}Approach}

    The sDropout model is an extension of Dropout, which keeps the same sampling process in fully-connected layers as Dropout does, but takes the thrown-away units as an extra inputs of the next layer. Then we discuss the rotation convolution techniques: rotate-pooling convolution (RPC) and flip-rotate-pooling convolution (FRPC). These two techniques extract rotation invariant features through the rotated and flipped filters in convolution layers combined with max-pooling.

\subsection{Split Dropout}

\begin{figure}[ht]
    \centering
    \includegraphics[width=1\linewidth]{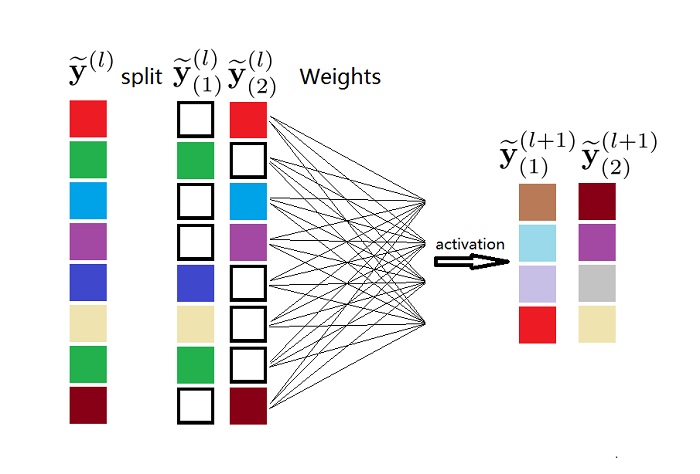}
    \caption{\label{fig:Split procedure}An example of a single sDropout layer. One output of layer $l$ is randomly sampled into two parts for layer ~$l+1$~, both taken for weight connection and activation. Finally, we get two outputs of layer $l+1$.(Recommend read in electronic copy) }
    \end{figure}

    Dropout was proposed by Hinton \etal~\cite{hinton2012improving} as a regularization to prevent the deep neural nets with a large number of parameters from overfitting. The key idea is to randomly drop units with probability $p$ from the network during training, which breaks the co-adapting among units. The survived units make up a ”thinned” network. A neural net with $n$ units can be seen as a collection of $2^n$ possible thinned neural nets. So training neural network with dropout can be seen as training a collection of $2^n$ thinned networks. At test, single neural net without dropout is used with the weights multiplied by drop probability $p$. In the experiments of this paper, $p$ is set to 0.5. This strategy achieves the effect of averaging exponentially many thinned models. As a result, the performance of neural nets with Dropout is significantly improved.\\
    \indent sDropout is an extension of Dropout. Similar to Dropout in the random sampling stage, sDropout randomly samples the input units, which breaks the co-adapting among units and thus prevents the model from overfitting. The key difference between them is that the units failing to survive in sampling stage are fed into an extra thinned network that shares the same weights with the previous sub-network, rather than discarded. Consequently, the inputs are split and trained in two sub-networks. The modification brings two benefits. On one hand, keeping all the units can guarantee that all theweights are trained in each iteration, while only half ($p$ is set to 0.5) of weights are updated in Dropout. As a result, this strategy will accelerate the convergence rate. On the other hand, averaging between the same network with different inputs help improve performance. Notably, sDropout inherits several advantages of Dropout: (1) it breaks co-adapting among units and thus prevent overfitting; (2) it approximately combining exponentially many different neural network architectures efficiently.\\
    \indent To illustrate our sDropout model, we consider a fully-connected layer $l$ of a neural network. Let ${\bf z}^{(l)}=[z_1,z_2,...,z_n]^T$ be the inputs of layer $l$ and $W^{(l)}$ (of size $d\times n$) be the weight parameters ( biases are included in $W^{(l)}$ with a corresponding fixed input of 1 for simplicity). The outputs from the layer $l$ are denoted by ${\bf y}^{(l)}=[y_1,y_2,...,y_d]^T$, computed by multiplying the input vector with the weights matrix followed by a non-linear activation function a. The feed-forward process of standard Dropout can be described as the follow:
    \begin{equation}
    \begin{array}{rcl}
    m_i^{(l)} &\sim& Bernoulli(p),\\
    {\bf\widetilde{y}}^{(l)} &=& {\bf m}^{(l)}\star{\bf y}^{(l)},\\
    z_i^{(l+1)} &=& {\bf w}_i^{(l+1)}{\bf\widetilde{y}}^{(l)},\\
    y_i^{(l+1)} &=& a(z_i^{(l+1)}).\\
    \end{array}
    \end{equation}
    \noindent where $\star$ denotes element-wise product and $\bf m $ is a binary vector of size $d$, each element is an independent Bernoulli random variable, $m_j\sim Bernoulli(p)$. Each element of a layer's outputs is kept with probability of $p$, otherwise is set to $0$ with probability of $(1-p)$. Thus, the outputs of layer $l$, i.e. ${\bf y}^l$, are multiplied with the random sample vector $\bf m$ and get a thinned outputs to be used as inputs to the next layer ~$l+1$~. Many experiments and theoretical analysis have indicated that Dropout improves the network's generalization ability and test performance. However, only $p$ part of the weights are renewed each iteration in this way and the training procedure is retard. \\
    \indent The proposed sDropout differs from Dropout in that the outputs of layer $l$, ${\bf y}^{(l)}$ are split into two parts, i.e. ~${\bf\widetilde{y}}_{(1)}^{(l)}$ and ~ ${\bf\widetilde{y}}_{(2)}^{(l)}$~, through a random sample vector $\bf m$. Then they are taken as two different inputs for the consequent weight connection and activation. This process picks up the dropped outputs $(\mathbf{1_e-m})\star{\bf y}^{(l)}$  as another inputs for layer $l+1$. The feed-forward process with sDropout is described in equation (\ref{equation:sDropout}), where $\bf 1_e$ denotes a vector of size $d$ with all elements being $1$. Figure~\ref{fig:Split procedure} illustrates the procedure of sDropout in one fully-connected layer.

    \begin{equation}
    \begin{array}{rcl}
    m_i^{(l)} &\sim& Bernoulli(p),\\
    {\bf\widetilde{y}}_{(1)}^{(l)} &=& {\bf m}^{(l)}\star{\bf y}^{(l)},\\
    {\bf\widetilde{y}}_{(2)}^{(l)} &=& {\bf (1_e-m)}^{(l)}\star{\bf y}^{(l)},\\
    z_{i,k}^{(l+1)} &=& {\bf w}_i^{(l+1)}{\bf\widetilde{y}}_{(k)}^{(l)}, k=1,2 ,\\
    y_{i,k}^{(l+1)} &=& a(z_{i,k}^{(l+1)}), k=1,2.
    \end{array}
    \label{equation:sDropout}
    \end{equation}

    Learning process is just the same as that of the standard Dropout, using stochastic gradient descent (SGD) manner. If we take the sampled network in standard Dropout network as sub-network, the network reconstructed by the dropped outputs can be seen as an extra-sub-network. Forward and backpropagation are done in sub-network and extra-sub-network. The cost function of sDropout model is computed by combining the sub-network and the extra-sub-network. Particularly, let $f(x;\theta, \bf m)$ denote the cost function of Dropout model, given data $x$, the parameters $\theta = {W}$ and the randomly-drawn sample mask $\bf m$. As Dropout is a stochastic regularization technique, the correct value of cost function is obtained by summing out over all possible sample masks $\bf m$. $L(W)$ and $L_s(W)$ denote the cost of Dropout and sDropout respectively. We can see that the cost of both are equal.

    \begin{equation}
    L(W)=E_{\mathbf{m}}[f(x;\theta, \mathbf{m})]= \sum_{\mathbf{m}} p(\mathbf{m})f(x;\theta,\mathbf{m}).
    \label{equation:cost of Dropout}
    \end{equation}
    \begin{equation}
    \begin{array}{rcl}
    L_s(W)&=&E_{\mathbf{m}}[(f(x;\theta, \mathbf{m})+f(x;\theta,\mathbf{1_e-m}))/2]\\
        &=&\sum_{\mathbf{m}} p(\mathbf{m})[(f(x;\theta,\mathbf{m})+f(x;\theta,\mathbf{1_e-m}))/2]\\
        &=&\sum_{\mathbf{m}} p(\mathbf{m})f(x;\theta,\mathbf{m})\\
        &=&L(W).
        \end{array}
    \label{equation:cost of sDropout}
    \end{equation}

    The gradient for each parameter is averaged over the training cases in each min-batch. In each iteration, all the parameters are updated, while the ratio of weights being updated in standard Dropout is $p$. We can expect that the convergence will be faster, as the weights have being updated more frequently. In addition, We get a better estimate of gradient through the split subnetworks and sDropout perform better than Dropout. sDropout still enjoy the benefit of regularization since inputs are randomly split into two groups of samples, which breaks the complex co-adapting among units.

    When it goes to test, as Dropout does, we use a single network without sDropout to approximate the average of the exponentially many thinned models. The weights used are scaled-down by multiplying drop probability $p$. Our sDropout makes it possible as standard Dropout does to train a huge number of different networks in a reasonable time.

     sDropout can be applied to multiple fully-connected layers in the same way of one-layer sDropout. Since every sDropout layer splits the network into two, in the case of applying sDropout in $n$ layers, there will be $2^n$ sub-networks. Forward and backpropagation are done in these $2^n$ subnetworks. Since fully-connected layers do not take much time compared to convolution layers, the added computing time is not much.
\subsection{Rotate-Pooling Convolution}

    The features automatically learned from images by the CNNs are meaningful~\cite{zeiler2013visualizing}. The convolution layers extract features from low-level to high level, and invariance becomes greater as layer becomes higher. Although data augmentation by adding translated, rotated or scaled data can bring invariance in some extent, it can not be restricted to upper layers and get high-level invariance. Thanks to the local receptive fields, weight-tying and pooling strategy of CNNs, translation invariance is hard-coded into the CNNs. However, this prevents the pooling units from capturing rotation invariance~\cite{ngiam2010tiled}. The work of Zeiler and Fergus ~\cite{zeiler2013visualizing} visualizes the features captured from the top and bottom layers in CNNs and validates that the network output is not invariant to rotation, expect for the object with rotational symmetry. However, rotation invariance is crucial to the feature extraction.

    To settle down the problem, we propose two rotational convolution techniques, i.e. rotate-pooling convolution (RPC) and flip-rotate-pooling convolution (FRPC). Similar to translation invariance encoded into the convolution layers, we introduce rotated filters using local receptive fields, weight-tying and pooling strategies to encode the rotation invariance into the convolution layers of CNNs. The convolution layers in original networks are replaced by RPC and the other stages are kept unchanged.

    We rotate the convolution filters and do max-pooling on the output feature maps. Respectively, the convolution filter is rotated 0, 45, 90,...,315 degrees in-plane and thus we get 8 filters sharing the same weights. Then we convolute the input feature map with these 8 filters and get 8 outputs, which can be viewed as a 8-channel map. Finally, we do max-pooling among the 8 channels to get a final feature map as one output of the layer. We call this strategy rotate-pooling convolution (RPC). CNNs with RPC enjoy the two benefits of $(1)$ being able to capture rotated features, and $(2)$ having no extra parameters to train. While RPC is only applied on last few layers, there will be only little extra time for computation.

\begin{figure}[ht]
    \centering
    \includegraphics[width=1\linewidth]{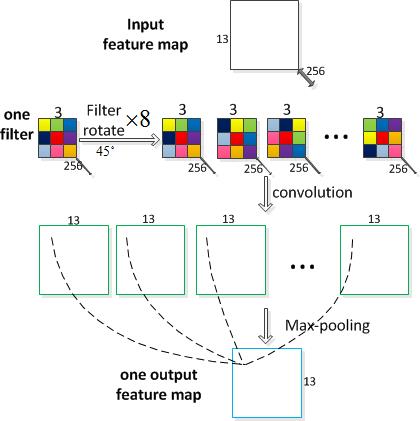}
    \caption{\label{fig:ZPSR} The rotate-pooling convolution in layer conv$3_1$. The first row is one input of layer conv$3_1$, a $13\times 13\times 256$ feature map. The second row shows the 8 rotated $3 \times 3\times 256$ filters sharing the same weights. By convoluting the input feature map with the 8 filters, we get 8 different feature maps in third row (in blue). Finally, through a max-pooling procedure which is element-wise over the 8 features maps, we get one output feature map of layer conv$3_1$.(Recommend read in electronic copy)}
    \end{figure}

    In this paper, we intend to extract rotation invariance in high-levels and thus we apply RPC in high-layers conv$3_1$, conv$3_2$, and conv$3_3$ of the CNNs architecture, which is described in section \ref{section:Implementation Details}. For example, Figure \ref{fig:ZPSR} shows the rotate-pooling convolution in layer conv$3_1$. To train the model, the feed-forward process is done as described previously and the backpropagation process is the same as the traditional max-pooling layer.  Note that our model with RPC does not change the Network architecture and training process, and can be easily applied to the convolution layers of other convolutional neural network based systems.\\
    \indent We let $r$ percentage of convolution filters undergo rotation, and the rest remains unchanged. We don't apply RPC to all the filters in each layer because the orientations of objects are sometimes useful for classification and recognition. Parameter $r$ is tunned on the training procedure. Our experiment result shows $r=50\%$ is better than $r=100\%$.
    .

    \subsection{Flip-Rotate-Pooling Convolution}

    Since we introduce rotation to the convolution filters, it i natural to consider about the flipped filters. In fact, the symmetric characters in images can be observed and human vision systems can recognize the rotated or flipped objects without any problem. Here, we modified the neurons in convolution layers by adding flipped filters. Similar to RPC, the selected filter is flipped up-down or right-left to produce one more filter sharing the same parameters. Then the input feature map is convoluted with 2 filters (1 original and 1 flipped) to produce a 2-channel feature map. Like what RPC does, a max-pooling across 2 channels is done and we get one output of the layer. We use FPRC to abbreviate the new convolution neuron Flip-Rotate-Pooling convolution in the following section.

    In the experiment, We also do not let all the filters undergo flip. The ratio of filters being flipped right-left and up-down is set by experiment adjustment and the selection of filters is random.

    \section{\label{section:Implementation Details}Implementation Details}

    We use the standard fully supervised convnet model of Zeiler and Fergus~\cite{zeiler2013visualizing} to validate our improvements on CNNs. Table~\ref{table:CNNs architecture} below shows the 8-layer architecture of the model we used in experiments. The model settings are the same as~\cite{zeiler2013visualizing} except that we use sDropout in the fully-connected layers fc$1$ \& fc$2$ and FRPC or RPC in conv$3_1$, conv$3_2$, \& conv$3_3$. In the rotational convolution layers, the percentage of filters to be rotated and flipped in each layer is set to $r=100\%$, $r=50\%$, or $r=25\%$. Note this process is the same for every sample.

    Our code is based on the Cuda-convnet package~\footnote{https://code.google.com/p/cuda-convnet/}. We adopt "data parallelism"~\cite{krizhevsky2014one} on the convolution layers. The GPUs are synchronized before the first fully-connected layer. We implement two GPUs, 64 image samples on each in every iteration.

    \begin{table}[h]
    \begin{center}
    \begin{tabular}{|c|c|}\hline
    layer & size \\
    \hline
    conv$1$    &$7\times7, 64,_{/2}$\\
    \hline
    pool$1$     &$3\times3, _{/2} $\\
    \hline
    conv$2$     &$5\times5, 256,_{/2}$\\
    \hline
    pool$2$     &$3\times3, _{/2}$\\
    \hline
    conv$3_1$    &$3\times3, 384$\\
    conv$3_2$    &$3\times3, 384$\\
    conv$3_3$    &$3\times3, 256$\\
    \hline
    pool$3$     &$3\times3,_{/2}$\\
    \hline
    fc$1$       &$4096$\\
    fc$2$       &$4096$\\
    fc$3$       &$1000$\\
    \hline
    \end{tabular}
    \end{center}
    \caption{\label{table:CNNs architecture}Architecture of 8 layer convnet model we use in this paper.   }
    \end{table}

\section{\label{section: experiment}Experiments}

    The proposed sDropout and the rotational convolution techniques are evaluated on the 1000-class ImageNet 2012~\cite{Russakovsky14}  classification task. The ImageNet Large Scale Visual Recognition Challenge~\cite{Russakovsky14} is a venue for evaluating the most effective image classification and recognition methods and CNNs have demonstrated a large step forward on ImageNet 2012 since~\cite{hinton2012improving, krizhevsky2012imagenet}. We use the provided 1.2 million training images for training. All the trained models are evaluated on the 50$k$ validation images with published labels and results are measured by top-1 and  top-5 error rates~\cite{Russakovsky14}.

    We use the following protocol for all experiments unless otherwise stated. For data preparation, each RGB image is preprocessed by resizing the small dimension to 256, cropping the center $256 \times 256$ region, subtracting the per-pixel mean and then using one randomly copped one-view subregion of size $224 \times 224$. Data augmentation is not used. In parameter setting, we use mini-batch SGD on batches of 128 images with learning rate of 0.2 and momentum parameter fixed at 0.9. All weights are initialized at 0.01 and biases are set to 1.

    We show error rate decreases of sDropout, RPC and FRPC on the CNNs model with activation function ReLU~\cite{nair2010rectified} or PReLU~\cite{he2015delving}. PReLU was proposed He \etal~\cite{he2015delving} which exhibits better performance than ReLU used in~\cite{zeiler2013visualizing} in deep neural networks. We adopt PRelu in our modifications to make the model more powerful. For convenience, let Z, ZP denote the model of Zeiler and Fergus~\cite{zeiler2013visualizing} with ReLU, PReLU respectively. Similarly, ZPS, ZPSR, ZPSRF for ZP model with \{ sDropout \}, \{ sDropout, RPC \} and \{ sDrpout, FRPC \} respectively. Overall, we observe a $1.18\%$ error rate decline (10-view, top-1) on model ZPSRF compared to ZP.

    \subsection{\label{section:sDropout VS Dropout experiment}sDropout v.s. Dropout}

\begin{figure}[ht]
\centering
\includegraphics[width=1\linewidth]{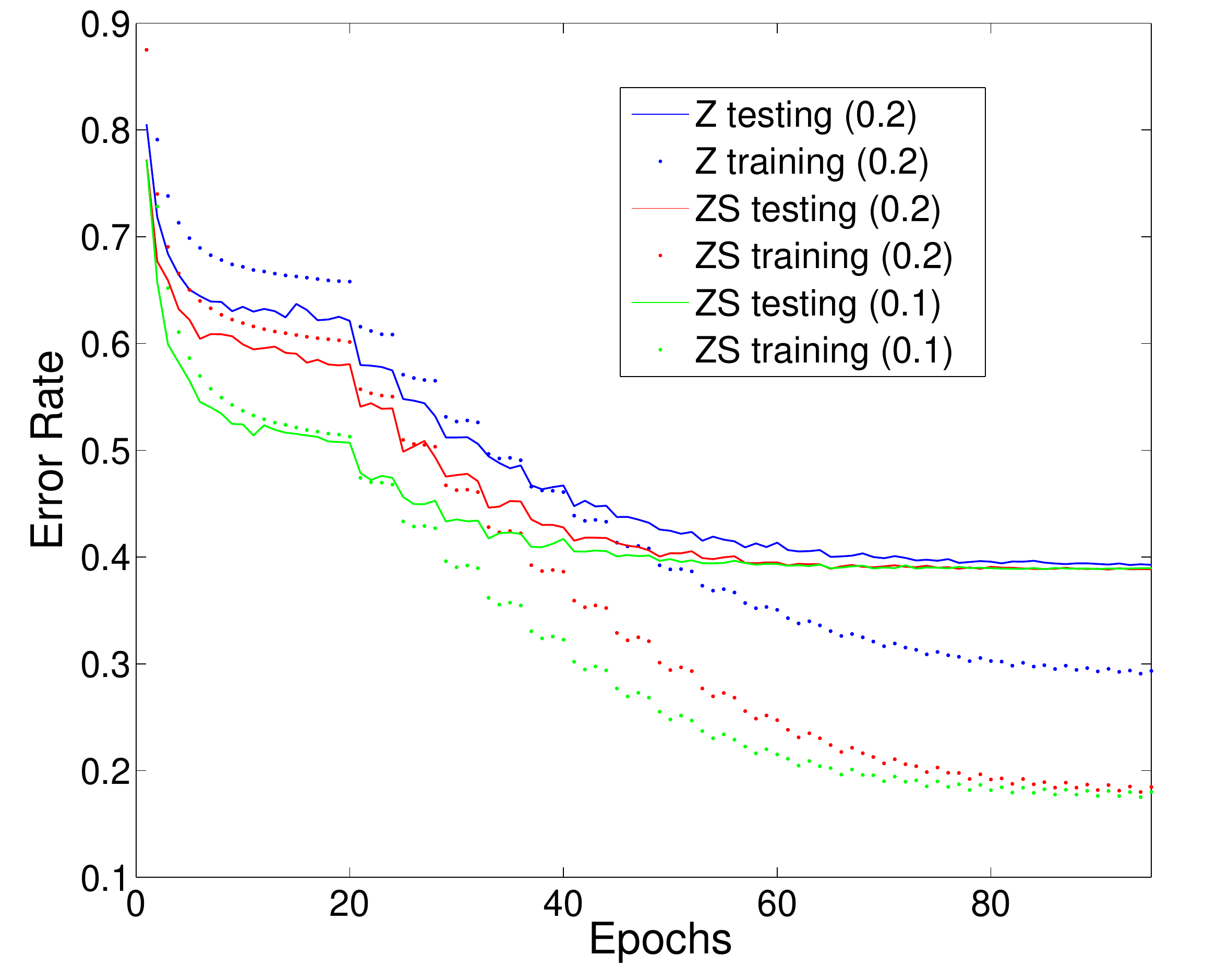}
\caption{\label{fig:ZP_ZPS_convergence}The convergence of model Z with Dropout and sDropout. The x-axis is the number of training and testing epochs. The y-axis is the 10-view top-1 error. The dot line corresponds to testing error rate on validation samples and solid line corresponds to training error rate. The model ZS with learning rate of $0.2$ ( in red ) has faster convergence than model Z with learning rate of $0.2$ on validation samples. In addition, model ZS has slightly lower error rate than model Z.(Recommend read in electronic copy)}
\end{figure}

\begin{figure}[ht]
\centering
\includegraphics[width=1\linewidth]{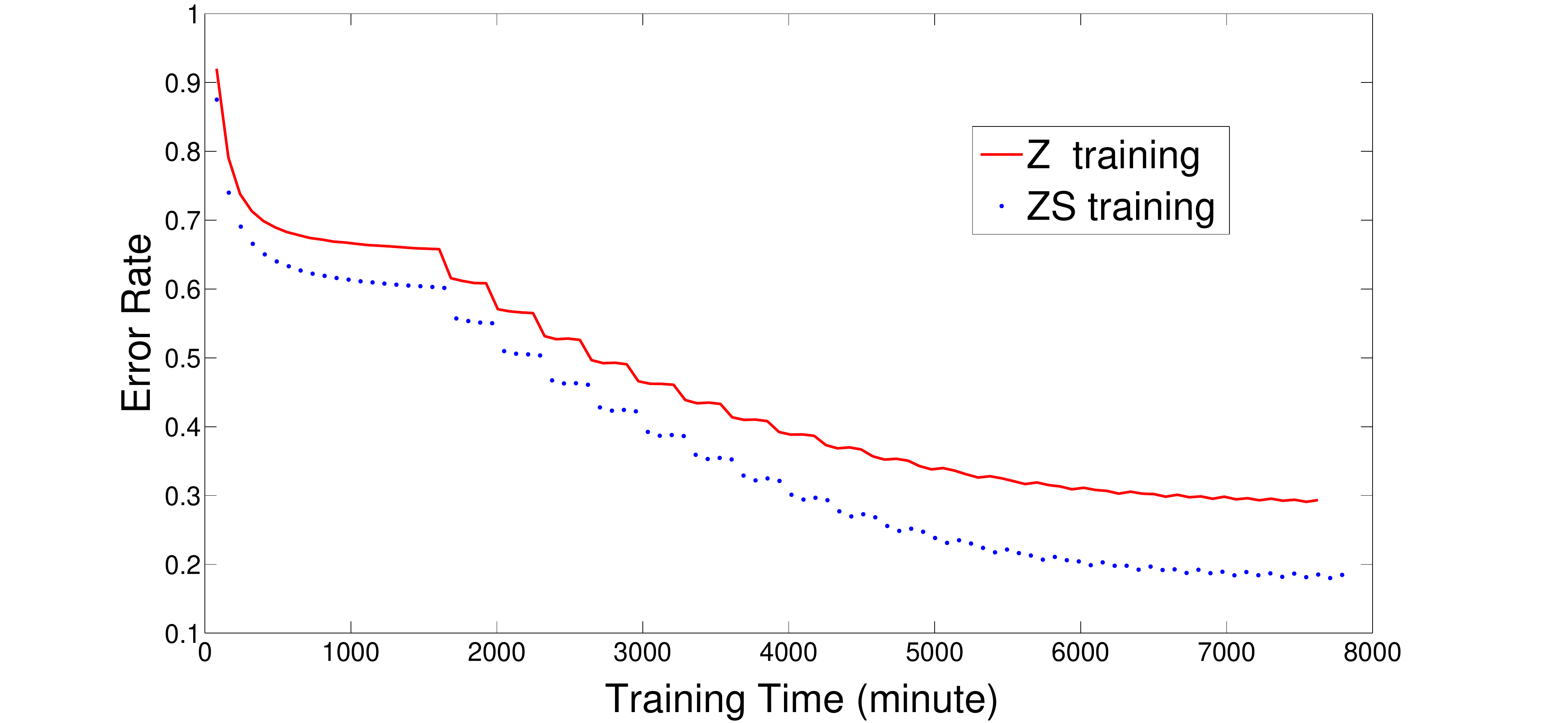}
\caption{\label{fig:ZP_ZPS_convergencetimeZ}The training time of model Z with Dropout and sDropout. The x-axis is training time. The y-axis is the 10-view top-1 error. The solid line (in red) corresponds to training error of model Z with Dropout and the dot line (in blue) corresponds to model Z with sDropout respectively. Model ZS takes little extra time compared to model Z.(Recommend read in electronic copy)}
\end{figure}

    We conducted comparisons between sDropout and standard Dropout using model Z and ZP. We apply Dropout and sDropout in the two fully-connected layers fc$_1$, fc$_2$. Figure~\ref{fig:ZP_ZPS_convergence} shows the convergence rate of model Z (with learning rate set to 0.1) and ZS (with learning rate set to 0.2) in training and testing. From the curves in the figure, we can find out that model ZS converges faster than model Z in both training and testing. In addition, the error rate of model ZS is 0.53\% lower than Z. On validation dataset, the top-1 and top-5 error rates of ZS (with learning rate set to 0.2) are 37.19\% and 15.69\%, while top-1 and top-5 error rates of Z (with learning rate set to 0.2) are 37.72\% and 15.81\% using 10-view testing. Despite the accuracy increment, sDropout takes little extra computing time. In figure \ref{fig:ZP_ZPS_convergencetimeZ}, the training time is compared between Z and ZS model.

\subsection{RPC \& FRPC}

\begin{figure}[ht]
    \centering
    \label{fig:oneclasspic}
    \includegraphics[width=1\linewidth]{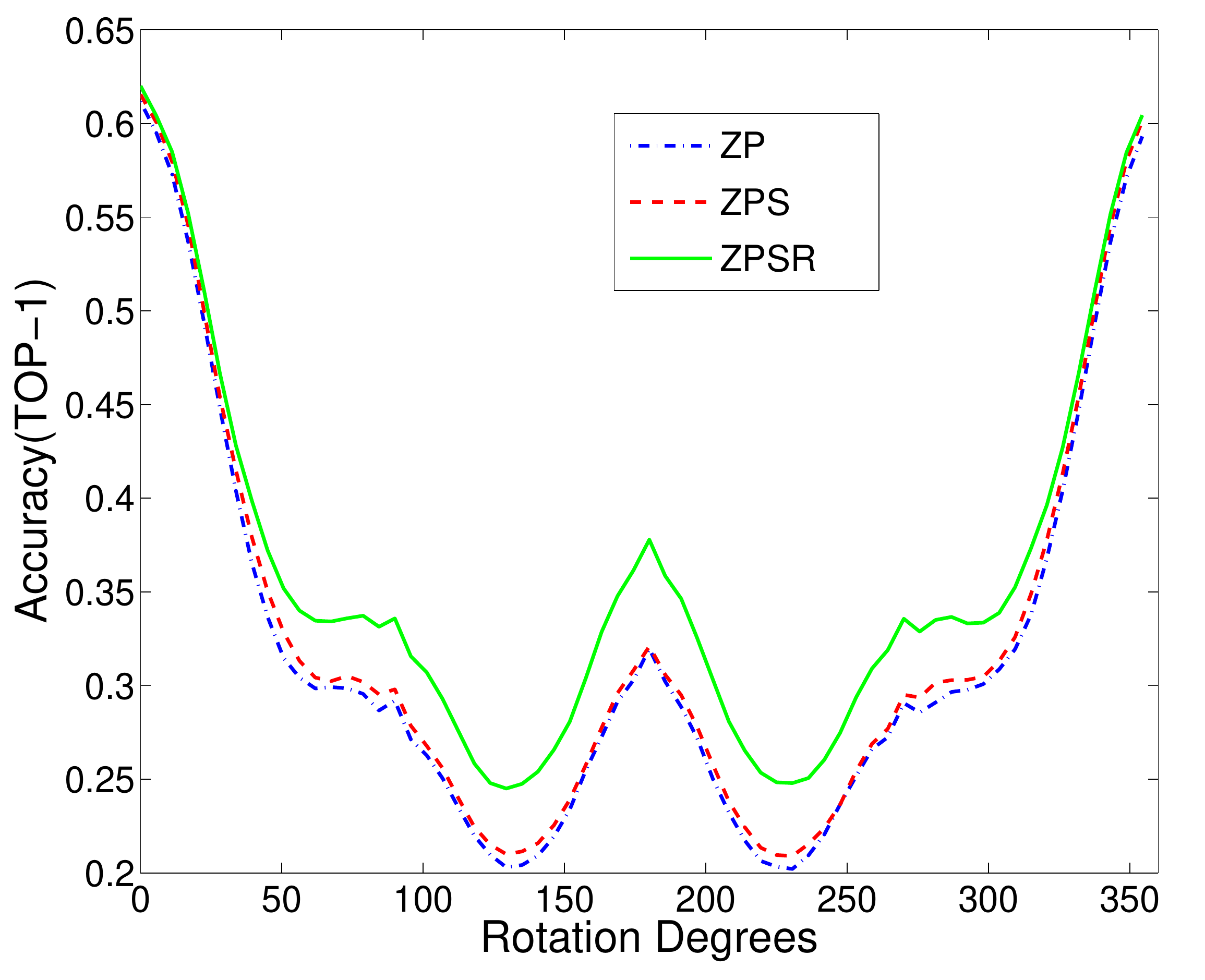}
    \caption{\label{Average rotate comparison}Comparisons between ZP, ZPS, and ZPSR on rotated testing dataset classification. The curves are average accuracy (TOP-1) on 50$k$ images over the rotation degrees. Model ZPS (in green) outperforms the other two models, especially at around 180\degree. (Recommend read in electronic copy)}
    \end{figure}

Here, we analyse the performance of model ZPSR on $50k$ ImageNet validation dataset. These images are rotated by various angles from 0\degree to 360\degree to get 64-views. We set the rotate percentage $r=100\%$ in this experiment, i.e all of the filters in conv$3_1$, conv$3_2$ and conv$3_3$ are rotated. In Figure~\ref{Average rotate comparison}, we see the average classification accuracy (top-1) of model ZP, ZPS and ZPSR over different rotation degrees. As the image rotation degree increases, the accuracy of the three models drops dramatically, which indicates that CNNs are susceptible to rotation transformation. Due to rotational symmetry of some images, peaks of the accuracy curve appear at rotation degrees of 90\degree, 180\degree and 270\degree. It is shown that model ZPSR always outperforms the other two, especially at around 180\degree. Besides, the average accuracy of model ZPS is slightly higher than ZP. It is validated that rotate-pooling convolution does improve the robustness of CNNs against rotation transform.

\begin{figure}[ht]
    \centering
    \subfigure{
    \label{fig:oneclasspic}
    \includegraphics[width=1\linewidth]{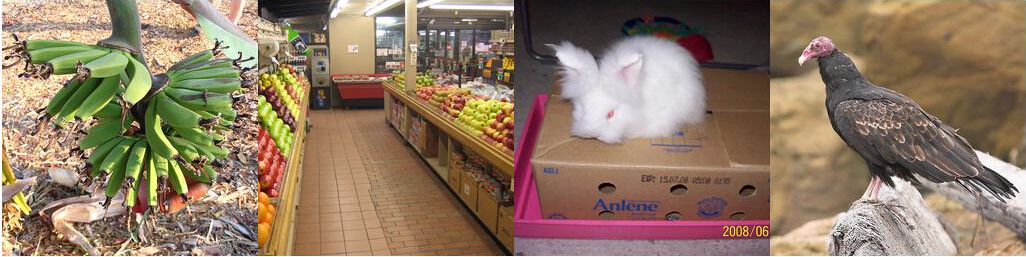}}
    \hspace{0in}\subfigure{
    \label{fig:oneclassfig}
    \includegraphics[width=1\linewidth]{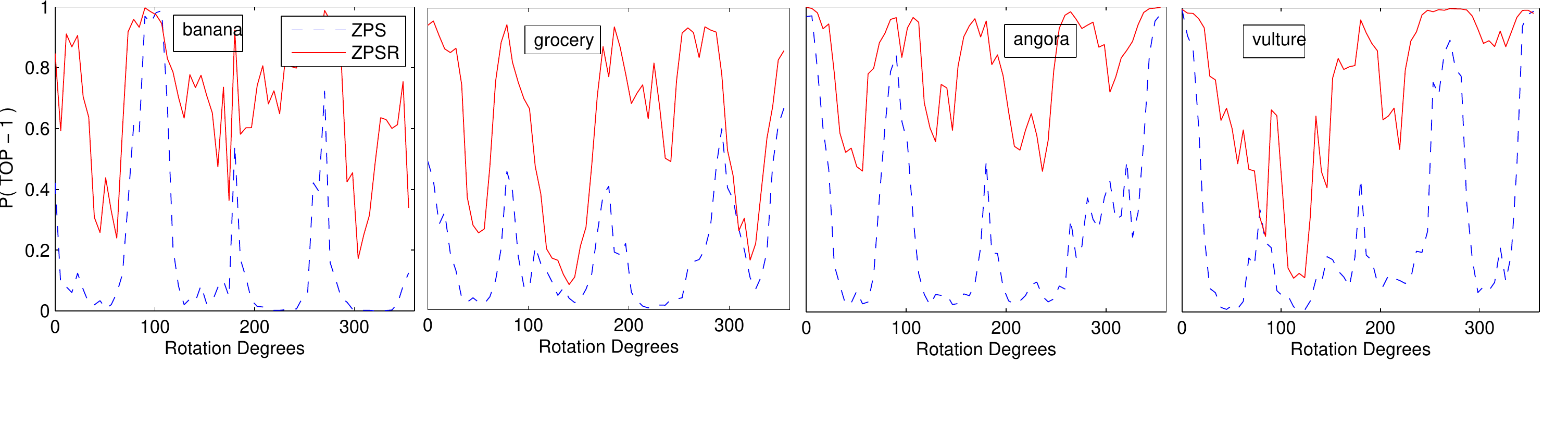}}
        \subfigure{
   \label{fig:twoclasspic}
    \includegraphics[width=1\linewidth]{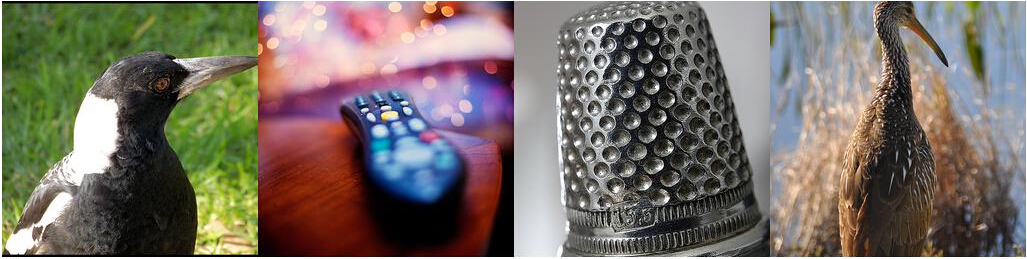}}
    \hspace{0in}\subfigure{
    \label{fig:twoclassfig}
    \includegraphics[width=1\linewidth]{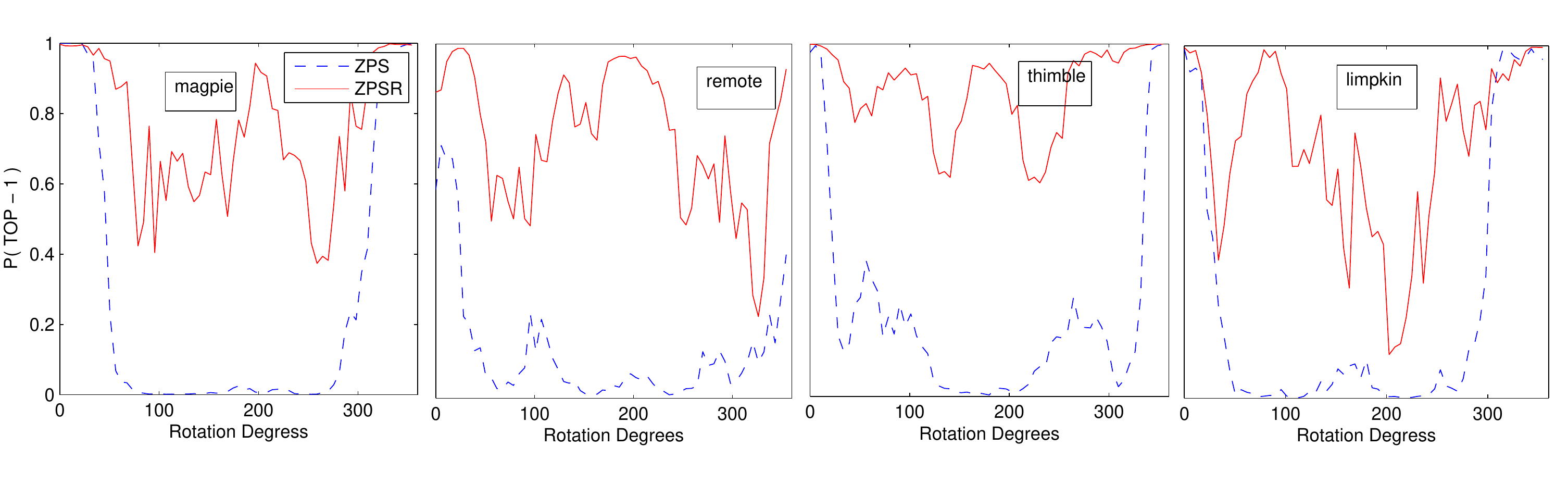}}
        \subfigure{
    \label{fig:threeclasspic}
    \includegraphics[width=1\linewidth]{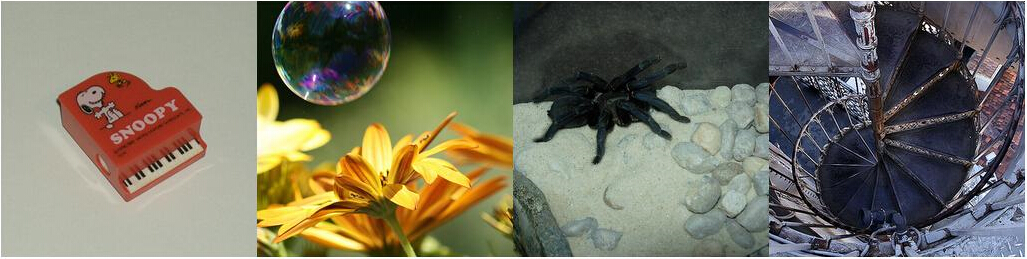}}
    \hspace{0in}\subfigure{
    \label{fig:threeclassfig}
    \includegraphics[width=1\linewidth]{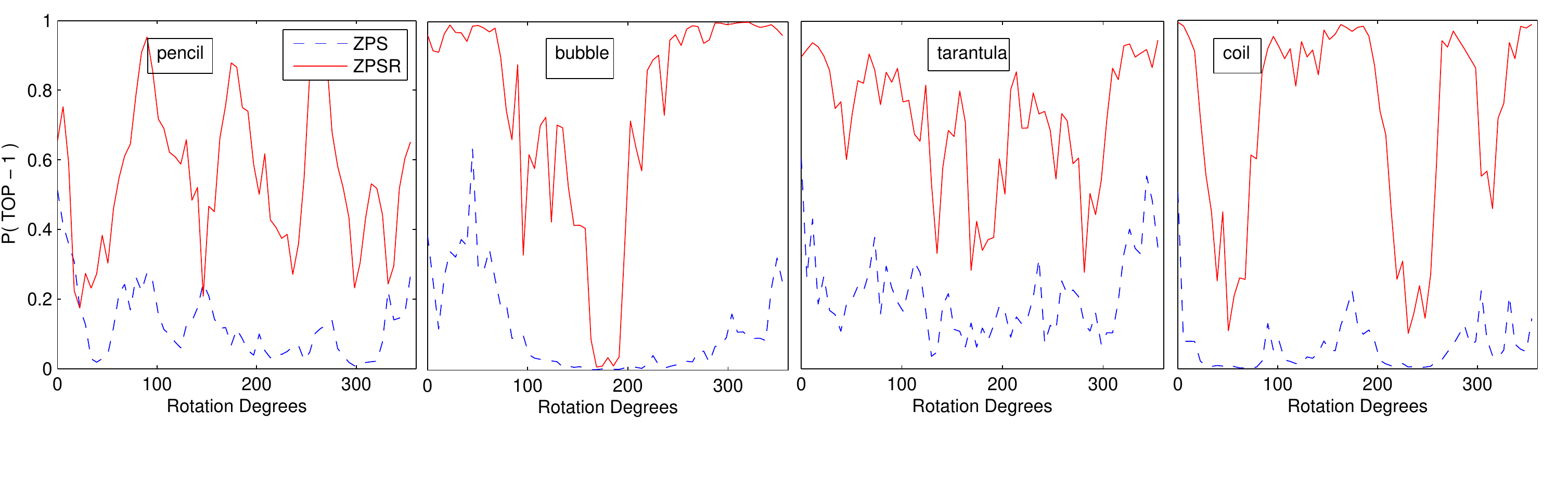}}
    \caption{\label{Analysis rotate}Analysis of rotation invariance within  model ZPS and ZPSR. Row 1, 3 \& 5 : example images undergoing rotation transformations. Row 2, 4 \& 6: the probability of the true label for each image against the rotation degree. (Recommend read in electronic copy)}
    \end{figure}

Since the classification accuracy is averaged by $50k$ images, we would like to explore the effect of RPC to different kinds of images. We pick some images in validation dataset that both models of ZPS and ZPSR can classify accurately. The classification results are compared through probability of true label against the rotation degrees.  Unsurprisingly, the effect of enhancement varies in different kinds of images. In general, probability of true label increases more or less by using Rotate-Pooling. To make it clear, we pick some typical images that have different features against rotations. As Figure~\ref{Analysis rotate} shows, images in top, middle and bottom are the typical images and their corresponding accuracy curves against different rotation degrees are shown below. As we can see, images in top row have some robustness to rotation transform at 90\degree, 180\degree, 270\degree in both model. The accuracy curve of model ZPS has four peaks at these degrees, while model ZPSR also has a better performance at other degrees. Images in middle are classified accurately only at small rotation degrees by model ZPS, while model ZPSR has high probability at most of the degrees. In the bottom, the images are classified with low probability at original degrees or others by model ZPS. For these images, model ZPSR has much higher probability at most of the degrees.

    \subsection{CNNs performance comparison}
    \begin{table*}[!ht]
    \begin{center}
    \begin{tabular}{|l||c|c|}\hline
    \cline{2-3}
    Error Rate \% & Val~ top-1 & Val~ top-5\\
    \hline
    Our replication of Zeiler and Fergus~\cite{zeiler2013visualizing}     &37.72 & 15.81 \\
    ZS                                                   &37.19 & 15.69 \\
    ZSRF (quarter)                                       &\bf{36.60} & \bf{15.28}\\
    ZP (Zeiler and Fergus~\cite{zeiler2013visualizing} with PReLU~\cite{he2015delving} )                                                 &37.15 & 15.62 \\
    ZPS                                                   &36.51 & 15.40 \\
    ZPSR(full)                                          &36.22 & 15.23\\
    ZPSR (half)                                           &36.08 & 15.10 \\
    ZPSRF (quarter)                                       &\bf{35.97} &\bf{15.06}  \\
    \hline
    \end{tabular}
    \end{center}
    \caption{\label{table:result compare}ImageNet 2012 classification error rates.}
    \end{table*}

    Finally, we compare the performance of models with sDropout or flip-rotate-pooling with different selected ratios on ImageNet 2012 classification~\cite{Russakovsky14}. These models sharing the same CNNs architecture (described in Table~\ref{table:CNNs architecture}) and parameters setting are (1) Z; (2) ZS; (3) ZSRF (quarter) ( rotate percentage $r=25\%$, flip percentage $r=25\%$);(4) ZP; (5) ZPS; (6) ZPSR(full) (rotate percentage $r=100\%$); (7) ZPSR(half) (rotate percentage $r=50\%$); (8) ZPSRF (quarter) (rotate percentage $r=25\%$, flip percentage $r=25\%$). Testing on 50k validation dataset, we use 10-view error rate of top-1 and top-5 to evaluate the classification result. The results are compared in Table~\ref{table:result compare} below. With sDropout, performance of Z and ZP is improved slightly from $37.72\%$ error rate to $37.19\%$ and $37.15\%$ to $36.51\%$. This improvement is obtained with almost no computational cost. Combined with sDropout, model ZPSRF with $25\%$ rotate and flip ratio has the lowest error rate among compared models on the testing data. sDropout and flip-rotate-pooling convolution technique totally bring 1.18\% accuracy increase to the ZP model.


\section{\label{section: discussion and conclusion}Discussion and Conclusion}

It is shown that introducing rotate-pooling convolution and flip-rotate-pooling convolution to the convolution layers is effective for improving classification performance of CNNs. Specifically, we find that using $25\%$ rotate and flip, FRPC and sDropout improves the model most. In addition, sDropout makes the network converges faster and slightly increase testing accuracy.

In this paper, we introduce Split Dropout, an extension of Dropout, to make the CNNs more efficient and effective on Image classification. In addition, we propose rotate-pooling convolution and flip-rotation convolution to make
the models robust to rotation transformations. These techniques bring considerable improvements on the task of classification, costing very little extra memory or computing time. It is worth mentioning that our formulation is inclusive of various techniques proposed recently such as multi-model averaging and dropconncet. The model possesses high potential to be applied in more sophisticated networks to achieve better performance.

{\small
\bibliographystyle{ieee}
\bibliography{FRPCsDropout}
}

\end{document}